# CLASSIFICATION OF POLAR-THERMAL EIGENFACES USING MULTILAYER PERCEPTRON FOR HUMAN FACE RECOGNITION

Mrinal Kanti Bhowmik
Department of Computer Science and Engineering
Tripura University , Suryamaninagar-799130
Agartala, India
email: mkb_cse@yahoo.co.in

Debotosh Bhattacharjee, Mita Nasipuri, Dipak Kumar Basu*,
Mahantapas Kundu
Department of Computer Science & Engineering
Jadavpur University - 700032
Kolkata, India
*AICTE Emeritus Fellow
email: debotosh@indiatimes.com
mita_nasipuri@gmail.com
dipakkbasu@gmail.com
mkundu@icse.jdvu.ac.in

*Abstract—* This paper presents a novel approach to handle the challenges of face recognition. In this work thermal face images are considered, which minimizes the affect of illumination changes and occlusion due to moustache, beards, adornments etc. The proposed approach registers the training and testing thermal face images in polar coordinate, which is capable to handle complicacies introduced by scaling and rotation. Polar images are projected into eigenspace and finally classified using a multi-layer perceptron. In the experiments we have used Object Tracking and Classification Beyond Visible Spectrum (OTCBVS) database benchmark thermal face images. Experimental results show that the proposed approach significantly improves the verification and identification performance and the success rate is 97.05%.

## I. INTRODUCTION

Face recognition has been an active research area for the last 20 years in the field of pattern recognition and computer vision, owing to its wide range of applications in commerce and law enforcement. It has many practical applications, such as bankcard identification, access control, mug shots searching, security monitoring, and surveillance systems [1] [2] [3]. Face recognition is used to identify one or more persons from still image or a video image sequence of a scene by comparing input images with faces stored in a database. Face recognition has the benefit of being a passive, non-intrusive system, which can verify personal identity without the consent of the concerned person or individual. Even though humans can detect and identify faces in a scene with little or no effort, building an automated system that accomplishes such objectives is very challenging. The challenges are even more profound when one considers the large variations in the visual stimulus due to illumination conditions, viewing directions or poses, facial expressions, aging, and disguises such as facial hair, glasses, or cosmetics. To address these issues several attempts have been made to increase the performance of face recognition.

Recognition performed by the human being can be simultaneously seen as a holistic and a feature analysis approach [4]. Automatic face recognition often favors only one of these aspects. Features used for description of faces are either biometric features of the face, like distances between parts of the face like nose and mouth, or more abstract features, like filter responses on a grid [5]. Template-based methods that attempt to match well- defined portions of the face (eye, mouth) belong to the analysis category [6], [7]. The Principal Component Approach (PCA) [8] [9] [10] describes images in terms of linear combinations of basis images, and thus represents a global holistic approach[11]. But results shows that PCA based approaches are poor in handling variations in scale, rotation, illumination and facial hair like beard and moustache. The main objective of this work is to improve the performance of the face recognition system subject to following variations:

**Scale invariance:** The same face can be presented to the system at different scales. This may happen due to the varying distance between the face and the camera. As this distance gets closer, the face image gets bigger.

**Pose invariance:** The same face can be presented to the system at different perspectives and orientations. For instance, pose of the face images of the same person may appear different due to rotation and tilting.

**Illumination invariance:** Face images of the same person can be taken under different illumination conditions because the position and the strength of the light source may change.

**Emotional expression and detail invariance:** Face images of the same person can differ when smiling or laughing and with some details such as dark glasses; beards or moustaches can be present.

Recently, researchers have investigated the use of thermal infrared face images for person identification to tackle illumination variation, facial hair, hairstyle etc. [12] [13] [14] [15] [16]. Log-polar transform [17] [18] [19] is applied to achieve rotation and scaling invariant images. In this paper, we present a novel approach to the problem of face recognition that realizes the full potential of the thermal IR band. In this work at first polar domain conversion of thermal images is done, after that using these transformed images eigenfaces are computed and finally those eigenfaces thus found are classified using a multilayer perceptron.

The organization of the rest of this paper is as follows. In section II, the overview of the system is discussed, in section III experimental results and discussions are given. Finally, section IV concludes this work.

## II. THE SYSTEM OVERVIEW

Here we present a technique for human face recognition. In this work we have used Object Tracking and Classification Beyond Visible Spectrum (OTCBVS) database benchmark thermal face images. Every face image is first converted into polar domain. These transformed images are separated into two groups namely training set and testing set. The eigenspace is computed using training images. All the training images and testing images are projected into the created eigenspace and named as polar thermal eigenfaces. Once these conversions are done the next task is to use a classifier to classify them. A multilayer perceptron is used for this purpose. The block diagram of the system is given in figure 1. In this figure dotted line indicates feedback from different steps to their previous steps to improve the efficiency of the system.

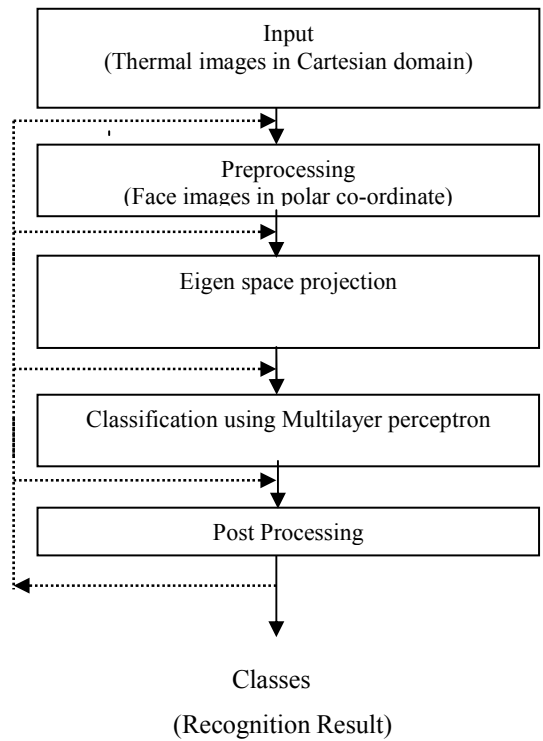

Figure 1. Block diagram of the system presented here.

### A. **Thermal Infrared Face Images**

Thermal infrared face images are formed as a map of the major blood vessels present in the face. Therefore, a face recognition system designed based on thermal infrared face images cannot be evaded or fooled by forgery, or disguise, as can occur using the visible spectrum for facial recognition. Compared to visual face-recognition systems this recognition system will be less vulnerable to varying conditions, such as head angle, expression, or lighting.

### B. **Log-polar transformation**

The log-polar transformation is used to get rid of the problems of rotation and scaling. This transformation maps thermal faces of size M x N into a new log-polar thermal face image of size $Z^q$ x $Z^q$, Z and q will be explained subsequently. Figure 2 shows that the rotation of faces in different angles appears just column shifted in polar domain. Scaling has got no effect if we use a fixed size for all the images in the polar domain, which can be easily understood from figure 3. The Log-polar transformation algorithm is described subsequently.

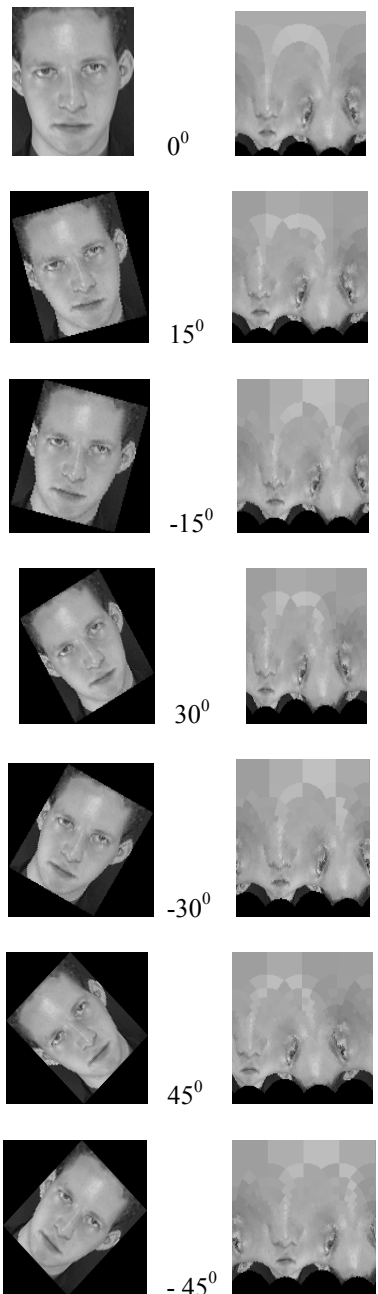

Figure 2: The Log-polar transformation for sample face image in rotation angles 0, +15, -15, +30, -30, +45, -45 degrees.

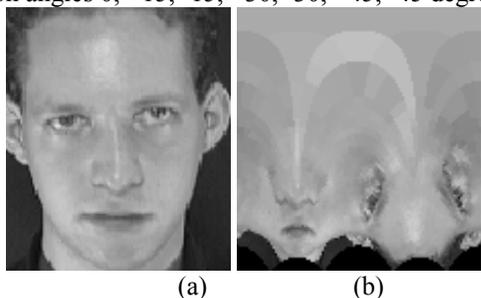

(a)          (b)

Figure 3: The Log-polar transformation of sample face image (first image of the figure 2.) in the scale of twice (a) and (b).

**Algorithm 1: Log-polar transformation**

Input: An image of size M × N in Cartesian coordinate space.
Output: An image of size $Z^q \times Z^q$ in Log-polar coordinate space.
Step 1: For given input image of size M × N, find the center (m, n) and radius (R) ensuring that the maximum number of pixels is included within the reference circle of the conversion. Center of the circle can be given as

$$m = \lfloor M/2 \rfloor, n = \lfloor N/2 \rfloor \quad (1)$$

Step 2: Compute polar images
The pixel in the input image $(x_i, y_i)$ will be the pixel at (r, $\theta$) position in the polar image, where

$$r = \sqrt{(x-m)^2 + (y-n)^2} \quad 0 \leq r \leq R \quad (2)$$

$$\theta = \tan^{-1}\left(\frac{y-n}{x-m}\right) \quad 0 \leq \theta \leq 360^0 \quad (3)$$

Step 3: Log-polar transform
Log-polar transform can be given as (p, $\theta$), where $p = \log_e r$.

Step 4: Resize the image obtained in step 3 into a square image of size $Z^q \times Z^q$, where $q = \lceil \log \frac{R}{Z} \rceil$.

During resizing we have used nearest neighbor interpolation as sharp boundaries are not very much useful feature in case of face recognition. Application of this algorithm in a thermal face image is shown in figure 4.

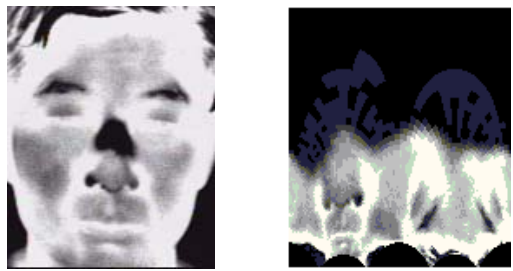

Figure 4. Example of conversion of thermal face in Cartesian space to log-polar coordinate space.

C. **Eigenfaces for Recognition**
In the language of information theory, we want to extract the relevant information in a face image, encode it as efficiently as possible, and compare one face coding with a database models encoded similarly. A simple approach is to extract the information contained in a face images, independent of any judgment of features, and use this information to encode and compare individual face images. In mathematical terms, we wish to find principal components [8] [9] [10] of the distribution of faces, or the eigenvectors of the covariance matrix of the set of face images. These eigenvectors can be thought of as set of features which together characterize the variations between face images. Each image location contributes more or less to each eigenvector, so that we can display the eigenvector as sort of ghostly face which we call an eigenface. Each face image in the training set can be presented exactly in terms of a linear combination of the eigenfaces. The number of a possible eigenfaces is equal to the number of face images in the training set. However the faces can also be approximated using only the "best" eigenfaces-those that have the largest eigenvalues, and which therefore account for the most variance within the set face images. The best U eigenfaces constitute a U-dimensional subspace, which may be called as "face space" of all possible images. Identifying images through eigenspace projection takes three basic steps. First the eigenspace must be created using training images. After that all those training images are projected into the eigenspace and call them eigenfaces. Train a classifier using these eigenfaces. Finally, the test images are identified by projecting them into the eigenspace and classifying them by the trained classifier.

**D. ANN using backpropagation with momentum:**
Neural networks, with their remarkable ability to derive meaning from complicated or imprecise data, can be used to extract patterns and detect trends that are too complex to be noticed by either humans or other computer techniques. A trained neural network can be thought of as an "expert" in the category of information it has been given to analyze. The Back propagation learning algorithm is one of the most historical developments in Neural Networks. It has reawakened the scientific and engineering community to the modeling and processing of many quantitative phenomena using neural networks. This learning algorithm is applied to multilayer feed forward networks consisting of processing elements with continuous differentiable activation functions. Such networks associated with the back propagation learning algorithm are also called back propagation networks.

III: EXPERIMENT RESULTS AND DISCUSSIONS

This work has been simulated using MATLAB 7. For comparison of results experiments are conducted for thermal. A thorough system performance investigation, which covers all conditions of human face recognition, has been conducted. They are face recognition under i) variations in size, ii) variations in lighting conditions, iii) variations in facial expressions, iv) variations in pose.
We first analyze the performance of our algorithm using OTCBVS database which is a standard benchmark thermal and visual face images for face recognition technologies.

**A. OTCBVS database**
Our experiments were perform on the face database which is Object Tracking and Classification Beyond Visible spectrum (OTCBVS) benchmark database contains a set of thermal and visual face images. There are 2000 images of visual and 2000 thermal images of 16 different persons. For some subject, the images were taken at different times which contain quite a high degree of variability in lighting, facial expression *(open / closed eyes, smiling /non smiling etc.)*, pose (Up right, frontal position etc.) and facial details (Glasses/ no Glasses). All the images were taken against a dark homogeneous background with the subjects in and upright, fontal position, with tolerance for some tilting and rotation of up to 20 degree. The variation in scale is up to about 10% all the images in the database.

**B. Classification of polar thermal eigenfaces using multilayer perceptron**
Out of total 2000 thermal images 1120 IMAGES are used as training set and rest 880 images are taken as testing images. All these thermal images are first transformed into polar domain. In this work a multilayer neural network with back propagation has been used. The learning algorithm error back propagation with momentum is used here. Momentum allows the network to respond not only to the local gradient, but also to recent trends in the error surface. We have used momentum to back propagation learning by making weight changes equal to the sum of a fraction of the last weight change and the new change suggested by the back propagation rule. The magnitude of the effect that the last weight change is allowed to have is mediated by a momentum constant, mc, which can be any number between 0 and 1. When the momentum constant is 0, a weight change is based solely on the gradient. When the momentum constant is 1, the new weight change is set to equal the last weight change and the gradient is simply ignored. The gradient is computed by summing the gradients calculated at each training example, and the weights and biases are only updated after all training examples have been presented.

The design of a neural network is not very easy. There are two major approaches to finalize the number of hidden layers and number of nodes in each of the hidden layers of a network [25]. The first method is called as pruning algorithm where the training process starts with a larger network so that an acceptable solution is found. After that, some hidden units are removed without degrading the performance of the network. The second method is called constructive approach, which starts with a small network and then grows with additional hidden nodes and layers to find acceptable solution. Constructive algorithm is straightforward and also finds a smaller network in comparison to pruning. In case of pruning it is difficult to find the initial network, which can classify the given inputs successfully. Therefore, it is better to use constructive approach. Moreover, in this work we have taken constructive approach [25], [26], keeping following two contradictory requirements:
    (i)    faster convergence of network

(ii) better classification performance

The Multi-layer perceptron network used here consists of five different layers. These are input, output and three hidden-layers viz. hidden layer-1, hidden layer-2 and hidden layer-3. Initially, experiments were conducted without any hidden layer, but the network did not converge. Therefore, hidden layers were introduced with an assumption that the given feature vectors are linearly non-separable. To decide about the number of neurons in hidden layer, several training experiments were conducted.

In this network we have used transig function which is a transfer function that produced output between 1 and -1. Learning rate and momentum constant used in this work are 0.02 and 0.9 respectively.

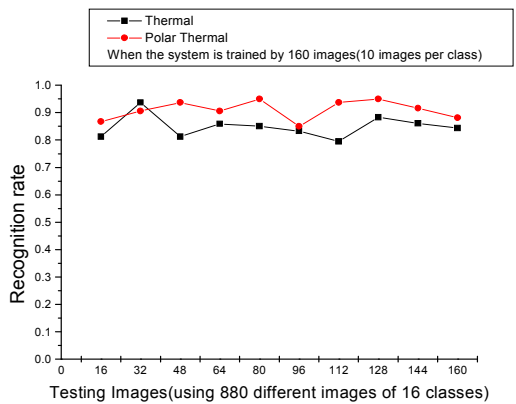

Figure 5. Comparative study of recognition rate of thermal and thermal polar face images.

In order to assess the effectiveness of thermal infrared images and polar transform, we compare face images from thermal spectra and their transforms in polar coordinate. Results obtained after applying the same procedure for thermal and thermal polar image are shown in figure 5. Here, experiments are conducted for different number of images during testing. Trend of false rejection error for different test images is depicted in figure 6.

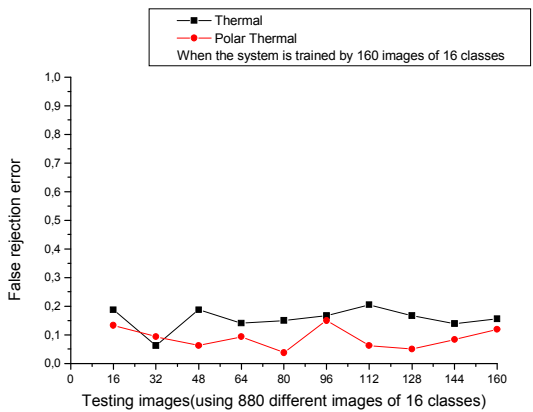

Figure 6. Comparative study of false rejection rate of thermal and thermal polar face images.

Comparison of recognition rates for the present method and other commonly referred methods are shown in Table 1 for a quick comparison. Although they use different face databases, i.e. other than OTCBVS face database, the present method can be compared favorably against other three recent face recognition methods.

**Table 1. Comparison of recognition rate with other Methods**

| Method | Recognition rate |
|---|---|
| Present Method | 97.05% |
| InfoGabor-GDA[21] | 95% |
| PCA [21] | 83.4% |
| Fusion of Thermal and Visual [15] | 90% |
| SQI [24] | 92% |
| Segmented Infrared Images via Bessel forms[13] | 90% |
| PCA for Visual indoor Probes[14] | 81.54% |
| Wavelet + RBF[22] | 96.3% |
| Wavelet Subband + Kernel associative Memory with XM2VTS database [23] | 84% |

IV: CONCLUSION

In this paper we have presented thermal face recognition results in varying lighting, facial expression, pose, and facial details. The scheme described here performs face recognition by combining the techniques of polar transformation, eigenspace projection, and classification using multilayer perceptron. Eigenspace projection has been advocated by the pattern recognition community for a long time for a broad range of applications. The efficiency of our scheme has been demonstrated on Object Tracking and Classification Beyond Visible spectrum (OTCBVS) benchmark database and recognition rate obtained is 97.05%. This scheme may be used for other type of pattern recognition and computer vision applications.

V. ACKNOWLEDGMENT

Authors are thankful to the "Centre for Microprocessor Application for Training Education and Research" and "Project on Storage Retrieval and Understanding of Video for Multimedia, at the Department of Computer Science and Engineering, Jadavpur University, Kolkata - 700 032 for providing the necessary infrastructural facilities for carrying out this work. First author also acknowledges the infrastructural support from Department of Computer Science and Engineering, Engineering faculty, Tripura University, Tripura, India.

**References**

[1] W. Zhao, R. Chellappa, A. Rosenfeld, and P. Phillips, " Face recognition: A literature survey", ACM Computer Survey, vol. 35, no. 4, pp. 399-458, December 2003.

[2] P. Jonathan Phillips, Alvin Martin, C.L.Wilson, Mark Przybocki, "An Introduction to Evaluating Biometric System", National Institute of Standards and Technology, page 56-62, February, 2000.

[3] A. Samal and P. A. Iyenger, "Automatic recognition and analysis of human faces and facial expressions : A survey", Pattern Recognition vol. 25, No. 1, 1992.

[4] R. Chellappa, C. L. Wilson, and S. Sirohey, "Human and machine recognition of faces: A survey," Proc. IEEE, vol. 83, pp. 705-740, 1995.

[5] M. Lades et al, "Distortion invariant object recognition in the dynamic link architecture," IEEE Trans. Comput., vol. 42, pp. 300-311, March 1993.

[6] R. Brunelli and T. Poggio, "Face Recognition: Features versus Templates", *IEEE Trans. Pattern Analysis and Machine Intelligence*, vol. 15, no. 10, pp. 1,042-1,052, 1993.

[7] A. Yuille, D. Cohen, and P. Hallinan, "Feature extraction from faces using deformable templates," in Proc. IEEE Comp. Soc. Conf. on Computer Vision and Pattern Recognition. New York: IEEE Comp. Soc. Press, 1989, pp. 104-109.

[8] M. Turk and A. Pentland, "Eigenfaces for recognition", Journal of Cognitive Neuro-science, March 1991. Vol 3, No-1, pp-71-86.

[9] L. Sirovich and M. Kirby, "A low-dimensional procedure for the characterization of human faces," J. Opt. Soc. Amer. A 4(3), pp. 519-524, 1987.

[10] Gottumukkal R, Asari VK (2004) "An improved face recognition technique based on modular PCA approach", Pattern Recognition Letters 25:429-436.

[11] B. Duc, S. Fischer, J. Bigun, "Face authentication with gabor information on deformable graphs", IEEE Tras. on Image Processing, vol. 8, no. 4, April 1999.

[12] Xin Chen, Patrick J.Flynn, Kevin W.Bowyer, "IR and Visible light face Recognition", University of NotreDame, USA, http://www.identix.com/products/.

[13] P. Buddharaju and I. Pavlidis and I Kakadiaris, "Face Recognition in the Thermal Infrared spectrum". Proceeding of the 2004 IEEE Computer Society Conference on Computer Vision and Pattern Recognition workshops (CVPRW'04).

[14] A. Socolinsky and A. Seinger, "Thermal face recognition in an operational scenario". Proceedings of the 2004 IEEE Computer Society Conference on Computer Vision and Pattern Recognition workshops (CVPRW'04).

[15] Singh, A. Gyaourva, G. Bebis and I. Pavlidis, "Infrared and Visible Image Fusion for Face Recognition", Proc. SPIE, vol.5404,pp.585-596, Aug.2004.

[16] P. Buddharaju, I. T. Pavlidis, P. Tsiamyrtzis, and M. Bazakos "Physiology-Based Face Recognition in the Thermal Infrared Spectrum", IEEE transactions on pattern analysis and machine intelligence, vol.29,no.4, April 2007.

[17] F. Smeraldi and J. Bigun, "Retinal vision applied to facial features detection and face authentication," Pattern Recognition Letters, vol. 23, pp. 463-475, 2002.

[18] M. Tistarelli and E. Grosso, "Active vision-based face recognition issues, applications and techniques," In Wechler, H. et al. (Editors), Face Recognition from Theory to Applications, vol. F-63, Springer, pp. 262-286, 1998.

[19] Richa Singh, Mayank Vatsa, Afzel Noore, and Sanjay K.Singh, "Age Transformation for improving Face Recognition Performance". Second International Conference,PReMI 2007 Kolkata,India,December 2007 Proceedings of Springer-Verlag Berlin Heidelberg 2007.page 576-583.

[20] J. Wang, K.N. Plataniotis and A.N. Venetsanopoulos, " Selecting discrimination eigenfaces for face recognition", Pattern Recognition Letters 26(2005), p.1470-482,

[21] Linlin Shen and Li Bai "Information Theory for Gabor Feature Selection for Face Recognition" Hindawi Publishing Corporation EURASIP Journal On Applied Signal Processing 2006, Article ID 30274, Page1-11.

[22] Meng Joo, Shiqian Wu, Juwei Lu, Hock Lye Toh, "Face Recognition with Radial Basis Function (RBF) Neural Networks", IEEE Transaction on Neural Networks, Vol.13, No.3, May 2002.

[23] Bai-Ling Zhang, Haihong Zhang, and Shuzhi Ge, "Face Recognition by Applying Wavelet Subband Representation and Kernel Associative Memory", IEEE Transaction on Neural Networks, Vol.15, No.1, May 2004.

[24] Haitao Wang, Stan Z Li, Yangsheng Wang, 'Face Recognition Under Varying Lighting Condition Using Self Quotient Image', Proceedings of the sixth IEEE International Conference on Automatic Face and Gesture Recognition(FGR'04).

[25] Kwok T , Yeung Y (1997) Objective Functions for Training New Hidden Units in Constructive Neural Network. IEEE Trans. on Neural Network, vol. 8, no. 5, September.

[26] Marchand M, Golea M, Rujan P (1990) A convergence theorem for sequential learning in two-layer perceptrons. Europhys. Lett. vol. 11 : 487-492.